\documentclass[letterpaper]{article} 
\usepackage[preprint]{aaai2027}  
\usepackage[hyphens]{url}  
\usepackage{graphicx} 
\usepackage{natbib}  
\usepackage{caption} 
\usepackage{todonotes}
\usepackage{booktabs}
\usepackage{array}
\usepackage{amsmath}
\usepackage{amssymb}
\usepackage{algorithm}
\usepackage{algpseudocode}
\usepackage{multirow}
\usepackage{cleveref}
\makeatletter
\renewcommand{\section}{\@startsection{section}{1}{\z@}%
  {-2.0ex plus -0.5ex minus -.2ex}%
  {3pt plus 2pt minus 1pt}%
  {\Large\bfseries\centering}}
\renewcommand{\subsubsection}{\@startsection{subsubsection}{3}{\z@}%
  {-6pt plus -2pt minus -1pt}%
  {-1em}%
  {\normalsize\bfseries}}
\makeatother

\title{FreeBalance: Pre-Routing Online MoE Load Balancing \\ via Residual Workload Prediction}

\author{
    Pengfei Chen\textsuperscript{\rm 1,2},
    Yize Wu\textsuperscript{\rm 1,2},
    Shouxu Kuang\textsuperscript{\rm 1,2},
    Ke Gao\textsuperscript{\rm 1,2},
    Ling Li\textsuperscript{\rm 1,2}
}
\affiliations{
    \textsuperscript{\rm 1}Institute of Software, Chinese Academy of Sciences, Beijing, China \\
    \textsuperscript{\rm 2}University of Chinese Academy of Sciences, Beijing, China
}

\begin{document}

\maketitle

\begin{abstract}

Load imbalance poses a major bottleneck to the efficiency of expert parallelism in distributed inference of Mixture-of-Experts (MoE) models. The most heavily loaded rank stalls global execution due to skewed routing distributions, directly increasing latency. While offline expert placement can alleviate persistent imbalance, practical multi-task serving workloads exhibit layer- and batch-dependent routing dynamics, making online load balancing indispensable. Existing approaches rely on routing statistics collected after each MoE router, requiring expert weight load or migration to begin only after routing decisions are available, consequently placing migration overhead on the inference critical path. In this work, we observe that online balancing can instead be largely overlapped with computation before target routing (e.g., attention), if routing distributions can be predicted accurately in advance. Therefore, we propose FreeBalance, a lossless online load-balancing framework that overlaps expert migration with preceding computation stages via residual workload prediction. FreeBalance leverages cross-layer similarities in hidden representations within the residual network to build a lightweight workload predictor. This enables proactive expert migration planning before routing decisions are available, creating substantial overlap between weight transfer and computation-heavy pre-routing stages. Furthermore, a cost model constrains the number of swaps to fully hide the synchronization overhead within the available window. Experiments across models and datasets show that FreeBalance reduces the max-to-mean rank load ratio by 32.8\% and end-to-end prefill latency by 13.1\%. Specifically, our method hides balancing overhead of an average of 5.1 experts per layer, which would otherwise account for ~8.5\% of the critical-path latency.

\end{abstract}

\begin{figure*}[t]
\centering
\includegraphics[width=0.9\textwidth]{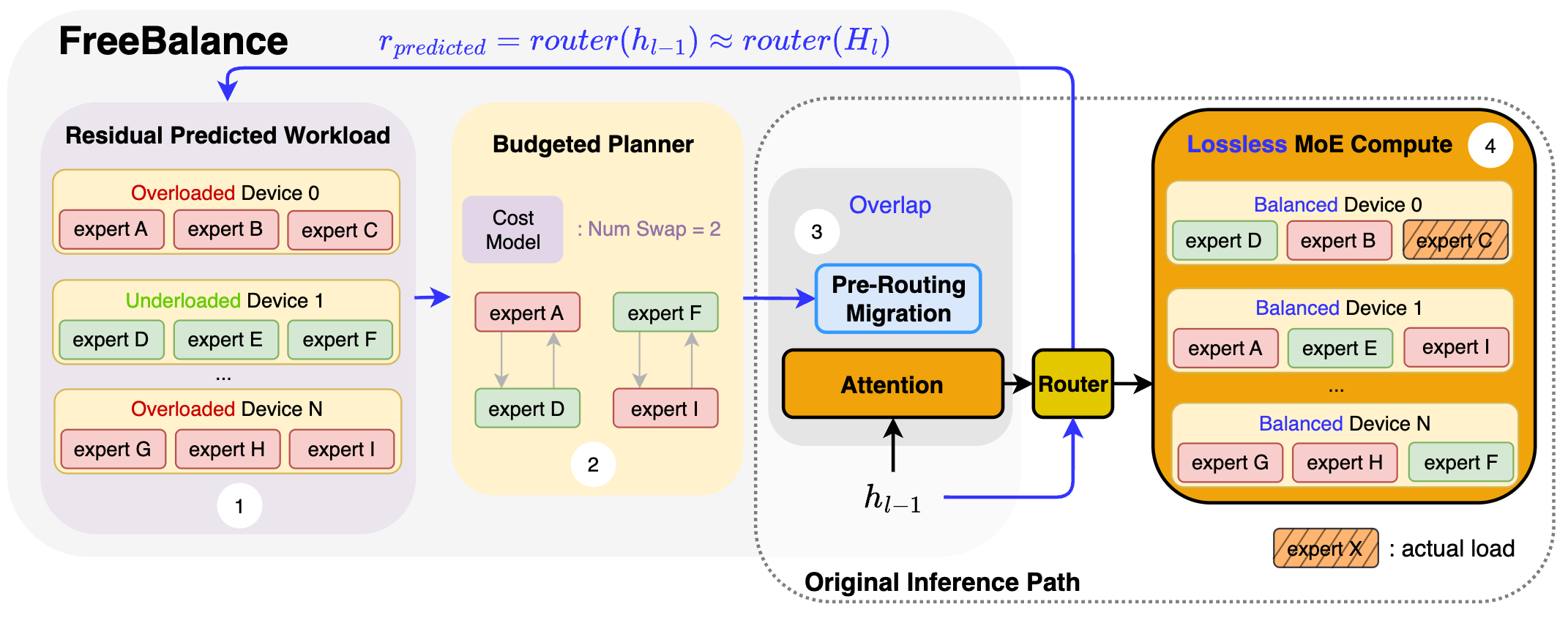}
\caption{Overview of FreeBalance and its four stages: \textcircled{1} residual workload prediction, \textcircled{2} budgeted expert-swap planning, \textcircled{3} expert-weight migration overlapped with pre-routing attention, and \textcircled{4} lossless MoE computation. The residual representation from layer $\ell{-}1$ predicts the expert workload of layer $\ell$ before its router executes. The prediction-guided expert placement is available before the target expert FFN executes. The blue line denotes the FreeBalance execution flow (ours), whereas the black line denotes the original inference flow.}
\label{fig:overlap}
\end{figure*}

\section{Introduction}

Mixture-of-Experts (MoE) architecture has become a dominant design choice for modern large language models (LLMs). It dramatically increases parameter capacity without proportionally increasing each-token computational cost \cite{deepseek2024v2,deepseek2024v3}. Specifically, each MoE layer activates only a small subset of experts for each input token through a lightweight routing mechanism, which effectively decouples performance capacity from inference latency \cite{shazeer2017outrageously}.

In distributed MoE inference, expert parallelism (EP) is commonly adopted for expert-stage acceleration. Experts of each MoE layer are sharded across multiple devices, allowing MoE computation to be executed in parallel. When tokens are routed to experts residing on different devices, the corresponding activation tensors are first dispatched to the corresponding devices via an all-to-all communication, then the results are combined after expert computation with a second all-to-all\cite{lepikhin2020gshard,rajbhandari2022deepspeed}. 

The effectiveness of EP is largely limited by load imbalance, which arises from uneven routing distribution across ranks. Despite the auxiliary load-balancing loss in the training process\cite{fedus2022switch}, inference-time routing distributions can still be highly skewed \cite{li2023accelerating}. Since both the all-to-all communication and subsequent expert computation require synchronization across the EP group, the overall execution latency is determined by the most heavily loaded rank. Consequently, lightly loaded devices must remain idle while waiting for stragglers to complete, resulting in reduced device utilization and degraded end-to-end performance.

Existing load balancing methods primarily rely on expert-device mapping modification, which changes the assignment of experts to physical devices to redistribute workloads more evenly. Such mappings can be determined offline based on historical routing statistics \cite{deepseek2025eplb}. However, practical LLM serving systems typically handle multi-task requests, where routing distributions can vary significantly across task types, and a static expert-device mapping cannot instantly adapt to dynamic workload changes, as illustrated in Figure~\ref{fig:routing_heatmaps}. In this case, online load balancing is therefore required, which dynamically adjusts expert-device assignments at runtime by migrating expert weights across ranks. However, the routing distribution required to guide expert placement is only available after the routing decision is made. Therefore, the available window for expert migration is restricted to the interval between routing completion and expert execution, which lies directly on the critical inference path and introduces additional latency overhead.

We identify the root cause of this problem as the sequential dependency between workload distribution acquisition and expert-map modification. Specifically, the routing stage can only begin after all preceding computations (e.g., attention computation of the current layer) are completed, and expert migration can only be initiated afterward based on the observed routing distribution. In this work, we observe that expert balancing does not necessarily need to wait for the current-layer router execution if the workload distribution can be predicted lightweightly and accurately before routing. Such pre-routing prediction enables expert migration to be overlapped with preceding computation stages, making it possible to effectively hide the balancing overhead. Furthermore, this approach is particularly beneficial for long-sequence inference, where attention computation often constitutes a significant portion of the critical latency.

Based on this insight, we propose FreeBalance, an online MoE load balancing framework (Figure~\ref{fig:overlap}) that performs pre-routing balancing through residual workload prediction. We observe that the routing distribution of an upcoming MoE layer can be well approximated before routing, via feeding the residual representation from the preceding layer into the corresponding router. This prediction is lightweight compared to the whole inference latency, and most importantly, it provides workload information before the actual routing stage. Based on the predicted workload, FreeBalance determines the online expert mapping adjustments and initiates weight migration immediately, so that the migration process can be overlapped with other pre-routing stages (primarily attention computations) and does not extend the critical path. Moreover, FreeBalance employs a cost model to constrain the amount of migrated experts according to the available overlapping opportunity, preventing overly aggressive adjustments from introducing additional latency. Notably, FreeBalance uses the predicted workload solely for online balancing, while the actual MoE execution still follows the original routing decisions, thereby achieving acceleration without affecting model outputs. Experiments show that FreeBalance reduces the max-to-mean rank load ratio by 32.8\%, reduces end-to-end latency by 13.1\%. Specifically, our method hides online balancing overhead of averagely 5.1 experts per layer, which would otherwise account for 8.5\% of the critical-path latency.

\section{Preliminary}

\subsection{Mixture-of-Expert Architecture}

An MoE layer contains a router and a set of feed-forward experts. For an input token representation $h_l$ of layer $l$, the router produces a score for each expert and selects the top-$k$ experts. 
\[
     W_l, I_l = \text{topk}(G(h_l)).
\]

Only the selected experts process the token, after which their outputs are weighted aggregated. 
\[
    MoE(h_l) = \sum_{i \in I_l} W_{l,i} * \text{expert}_i(h_l),
\]

where $W_{l,i}$ denotes the routing weight for expert $i$. The aggregated result is then forwarded as the output of the MoE block (added to $h_l$ along with the output of the attention block), such that
\begin{equation}
    h_{l+1} = h_l + \text{Attention}(h_l) + \text{MoE}(h_l+\text{Attention}(h_l))
\label{eq:resnet}
\end{equation}


\subsection{Expert Parallelism}
\label{sec:expert_parallelism}

Under EP, each expert is placed onto one or more specific devices according to an expert–device mapping (typically an even distribution). Each device hosts a subset of experts and processes tokens routed to them. When tokens are assigned to experts that reside on remote devices, an all-to-all communication is required to dispatch token representations to the corresponding devices. After computation, the expert outputs are communicated back to the original devices via a second all-to-all, where a final weighted aggregation is performed. 

\subsection{Load Imbalance}

Load imbalance poses a severe challenge under EP, arising from the skewed token routing distribution across EP ranks. When hot experts are colocated on the same rank, expert-level routing skew translates into inter-rank workload imbalance. As MoE execution requires global synchronization operations of result-gathering communication and expert-output aggregation, the most heavily loaded rank determines the critical path, forcing less-loaded ranks to wait in idle.



\begin{figure}[t]
\centering
\includegraphics[width=\columnwidth]{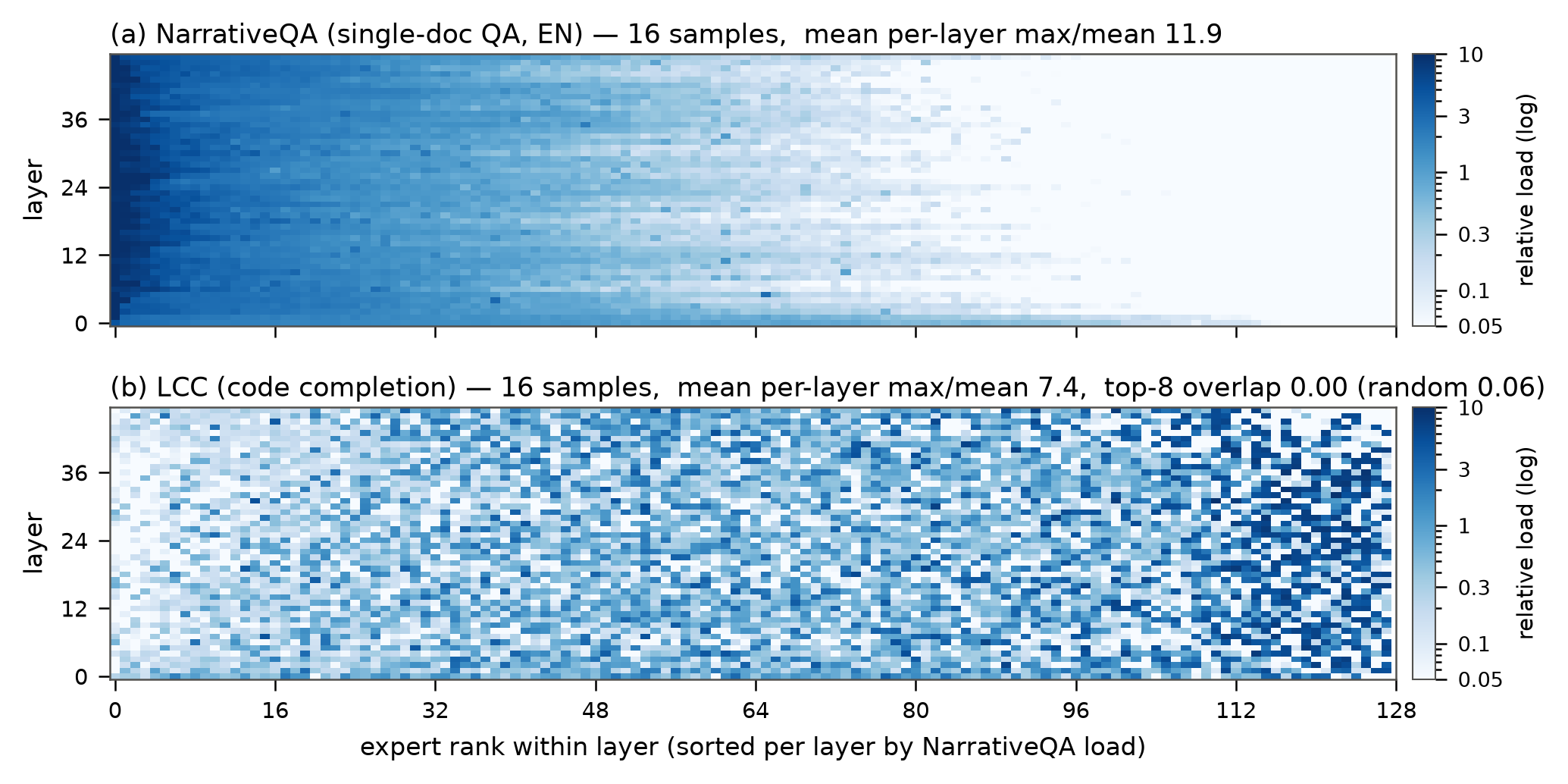}
\caption{Per-layer relative expert loads for NarrativeQA and LCC. Experts are sorted independently at each layer by their NarrativeQA load, and the same ordering is used for LCC to expose the workload-dependent shift in expert popularity.}
\label{fig:routing_heatmaps}
\end{figure}

Furthermore, in practical multi-task serving scenario, routing skew varies significantly across task types rather than remaining fixed. \Cref{fig:routing_heatmaps} presents the heatmap of workload distributions of two representative tasks on Qwen3-30B. As demonstrated, workload distributions differ substantially across tasks, with only 4.1\% of expert-activation overlap across all layers.




\subsection{Load Balancing}
\label{sec:load_balancing}




Offline load balancing typically decides expert placement for more balanced workload distributions from historical routing statistics. Despite its effectiveness, in multi-task scenarios where workload distributions vary significantly across tasks (as discussed in \cref{sec:expert_parallelism}), a fixed placement strategy may become suboptimal for the varying workload. In this case, workload-aware online balancing during inference is essential, which dynamically adapts expert-device assignments by migrating expert weights across ranks.

The key challenge of online load balancing lies in the additional overhead of expert migration. Since placement decisions can only be made after the routing distribution is computed, the available window for expert migration is limited to a narrow interval between routing and MoE computation, which falls directly on the critical inference path. Consequently, the introduced additional overhead can largely diminish the benefit of improved load balance.

\section{Method}

\subsection{Motivation and Overview}

As discussed above, the primary bottleneck of online load balancing is the migration overhead incurred on the critical inference path. Since expert placement decisions depend on the routing distribution, expert migration cannot begin until the target layer's router has completed, i.e., after the preceding attention computation. Consequently, only a narrow window remains for planning and expert-weights migration before MoE computation.


We observe that a substantial overlap opportunity naturally exists on the critical path. The pre-routing stages (e.g., attention computation) are computation-heavy, especially for long-sequence workloads, providing a sufficiently long window to hide expert migration latency. However, this opportunity remains unexploited because routing statistics become available only after the router executes, and migration planning can begin only afterward. Consequently, the entire pre-routing computation stage is unavailable for overlapping expert migration.

FreeBalance eliminates this limitation by breaking the sequential dependency between routing and migration planning. The key insight is that if the workload distribution can be accurately predicted before routing execution, migration planning no longer needs to wait for the actual routing results. Expert weight migration can therefore be scheduled proactively and overlapped with preceding computation stages. Therefore, we propose residual workload prediction for lightweight and accurate online workload prediction before the actual routing execution. Furthermore, to maximize the latency benefit while limiting migration overhead, FreeBalance employs a cost model to bound the migration budget and control migration planning. Furthermore, migration plans are constructed using pairwise expert swaps, ensuring that every rank preserves the same number of expert slots and memory footprint.

Notably, FreeBalance preserves the original routing decisions during MoE execution, ensuring lossless inference. The predicted workload distribution is used solely for expert placement decisions and weight transfer, while the target router determines the final token-to-expert assignments (rather than skipped). The final routing results may alter after the expert migration, yet the overall correctness is totally guaranteed.

In conclusion, for each MoE layer $\ell$, FreeBalance performs four stages:
\begin{enumerate}
    \item \textbf{Residual workload prediction.} A lightweight prediction module estimates the token counts of all experts before routing.
    \item \textbf{Budgeted Expert-Swap Planning.} A cost model determines the migration budget, and pairwise expert swaps are selected within the available overlap budget.
    \item \textbf{Weight Transfer.} Expert weights are exchanged across EP ranks, with the migration latency overlapped with computation-heavy pre-routing stages (mainly the attention stage).
    \item \textbf{Lossless MoE computation.} The target router computes the exact routing decisions, and MoE execution proceeds using the migrated expert placement, achieving a more balanced workload without altering the original routing behavior.
\end{enumerate}

\subsection{Residual Workload Prediction}

\begin{figure}[t]
\centering
\includegraphics[width=0.75\columnwidth]{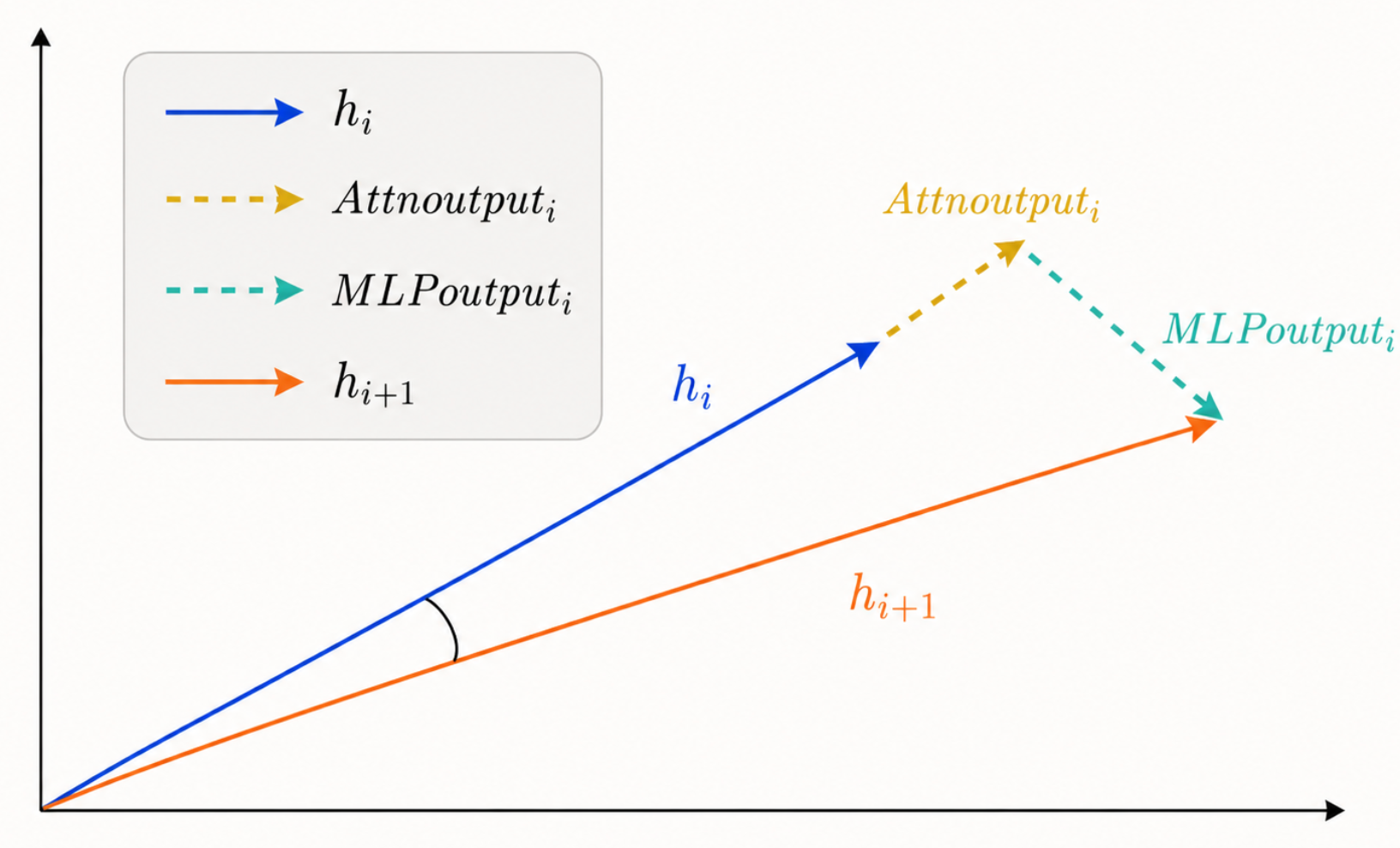}
\caption{Geometric intuition for cross-layer residual similarity. The attention and MoE updates cause only a small directional change from $h_{\ell-1}$ to $h_{\ell}$.}
\label{fig:residual_similarity}
\end{figure}

Residual connections make the hidden representations of adjacent layers highly similar~\cite{lee2024infinigen,liu2023dejavu,wu2025easyspec}. As illustrated in \cref{fig:residual_similarity} and formalized in \cref{eq:resnet}, $h_{\ell}$ is obtained by adding the attention and MoE updates to $h_{\ell-1}$. Because this residual update is small relative to the hidden-state magnitude, the angle between $h_{\ell-1}$ and $h_{\ell}$ is small, and their cosine similarity is close to 1. FreeBalance exploits this cross-layer similarity for lightweight and accurate online workload prediction. Since $h_{\ell-1}$ is available before the subsequent attention and routing stages, expert-placement planning and weight migration can begin at pre-routing time.

Migration must begin before the target layer routes its tokens, so the final router output is available too late to guide its own placement for that same layer. FreeBalance instead predicts the workload of $layer_\ell$ from $h_{\ell-1}$, the output hidden states of $layer_{\ell-1}$ and therefore the residual-stream input to $layer_\ell$. This state is available before the target layer performs attention and normal routing. Because consecutive layers process the same token sequence, $h_{\ell-1}$ retains task- and token-specific information that is useful for anticipating the next router's aggregate demand. The prediction guides only physical expert placement and never replaces model's normal routing pass.

Let $h_{\ell-1}\in\mathbb{R}^{N\times d}$ denote the output of layer $\ell-1$ for a prefill batch of $N$ tokens. It is also the residual hidden state entering target layer$\ell$ before attention. Let $H_{\ell}\in\mathbb{R}^{N\times d}$ denote the target router's normal input after the target layer's pre-routing transformations. FreeBalance introduces no standalone workload predictor; instead, it invokes the same frozen router $g_{\ell}$ twice:

\begin{align}
\widehat{G}_{\ell}&=g_{\ell}(h_{\ell-1}), &&\text{early pre-routing logits},\\
G_{\ell}&=g_{\ell}(H_{\ell}), &&\text{final routing logits}.
\end{align}
For a linear router, both invocations share the original routing function
\begin{equation}
g_{\ell}(X)=XW_{\ell}+\mathbf{1}b_{\ell}^{\top},
\end{equation}
where $W_{\ell}\in\mathbb{R}^{d\times E}$ and $b_{\ell}\in\mathbb{R}^{E}$ are the original router parameters. Applying the original top-$k$ rule to $\widehat{G}_{\ell}$ yields estimated assignments $\widehat{\mathcal{T}}_{\ell}(i)$ for token $i$. Aggregating these assignments gives the predicted expert workload
\begin{equation}
\widehat{n}_{\ell,e}=\sum_{i=1}^{N}
\mathbf{1}\!\left[e\in\widehat{\mathcal{T}}_{\ell}(i)\right].
\end{equation}

Because $h_{\ell-1}$ precedes the target layer's attention whereas $H_{\ell}$ follows its pre-routing transformations, the early assignments are estimates rather than final model decisions. Both invocations nevertheless use the same router parameters. Consequently, FreeBalance adds no predictor weights, training procedure, or checkpoint state. The early invocation produces only an $E$-element workload vector for placement planning; the normal invocation on $H_{\ell}$ remains unchanged and exclusively determines the final token-to-expert assignments.

The choice of $h_{\ell-1}$ balances timeliness and prediction fidelity. It incorporates all task- and token-dependent transformations through layer $\ell-1$, yet is available before attention in layer $\ell$. In contrast, $H_{\ell}$ is closer to the target routing decision but becomes available only when the normal router is about to execute. Using $h_{\ell-1}$ therefore creates an attention-length migration window while retaining sufficient information to estimate which experts will be hot in the target layer.

The early workload estimate is used only to plan physical expert placement, while the normal router invocation remains authoritative and determines which experts process each token. An empty swap plan retains the current placement. Therefore, the rebalancing procedure preserves the model's routing semantics and baseline execution.

\begin{algorithm}[t]
\small
\caption{FreeBalance for Target Layer $\ell$}
\label{alg:paired}
\algrenewcommand\algorithmicrequire{\textbf{Input:}}
\algrenewcommand\algorithmicensure{\textbf{Output:}}
\begin{algorithmic}[1]
\Require $h_{\ell-1}$, $g_{\ell}$, $\pi_{\ell}$, migration budget $B_{\ell}$, and threshold $\tau$
\Ensure Budget-feasible swap plan $\mathcal{S}_{\ell}$ and updated placement $\pi^{'}_{\ell}$

\State Predict next-layer expert workloads using $g_{\ell}(h_{\ell-1})$
\State Aggregate expert workloads into rank loads under $\pi_{\ell}$
\State $\mathcal{S}_{\ell}\gets\varnothing$
\While{rank-pair imbalance exceeds $\tau$ and budget remains}
    \State Select a beneficial swap between overloaded and underloaded ranks
    \If{no such swap exists}
        \State \textbf{break}
    \EndIf
\EndWhile
\State Migrate the selected experts during target-layer attention
\State Commit the placement before the target MoE stage begins
\State Execute the target MoE with the normal router decisions
\State \Return $\mathcal{S}_{\ell}$ and $\pi^{'}_{\ell}$
\end{algorithmic}
\end{algorithm}

\subsection{Budgeted Expert-Swap Planning}

Given the current batch's expert workload obtained by residual workload pre-routing, the planner seeks a placement that lowers the predicted maximum rank load without scheduling more communication than target-layer attention can hide. It must also produce the same result on every rank. We formulate planning as a budgeted optimization over pairwise swaps and use a deterministic greedy procedure that is lightweight enough to run at layer granularity.

Consider expert $e_a$ on rank $a$ and expert $e_b$ on rank $b$. Exchanging them changes only the loads of these two ranks:
\begin{align}
L'_{a}&=L_a-\widehat{n}_{e_a}+\widehat{n}_{e_b},\\
L'_{b}&=L_b-\widehat{n}_{e_b}+\widehat{n}_{e_a}.
\end{align}
We score a candidate by its reduction in a load objective
\begin{equation}
\Phi(L)=\max_r L_r+\gamma\sum_r(L_r-\overline{L})^2,
\end{equation}
where the first term targets the straggler rank and the second discourages moving the bottleneck to a different rank. The predicted benefit is $\Delta(e_a,e_b)=\Phi(L)-\Phi(L')$. Candidates with non-positive benefit are discarded.

Pairwise exchange has three practical advantages. It preserves the number of experts per rank, requires only bounded staging memory, and makes rollback straightforward because the old and new owners are known. It also avoids replication, which would require the router to divide assignments among multiple copies and would change the dispatch policy.


The usable migration budget is derived from a profiled attention time and the topology between the participating ranks. For a swap $s$, the planner estimates
\begin{equation}
C(s)=\alpha_{a,b}+\frac{S(e_a)+S(e_b)}{\beta_{a,b}},
\end{equation}
where $S(e)$ is the transferred representation of expert $e$, $\beta_{a,b}$ is the measured point-to-point bandwidth, and $\alpha_{a,b}$ captures launch and protocol overhead. The parameters are calibrated during runtime initialization and may distinguish intra-node and inter-node links.

For a fixed input shape, layers that use the same attention mechanism have nearly identical attention latency. FreeBalance therefore profiles attention once per attention type and reuses the measurement for subsequent layers of that type. Let $r(\ell)$ denote the first layer that uses the same attention mechanism as target layer $\ell$. In a homogeneous model, $r(\ell)=0$, so the layer-0 attention time is reused throughout the model. For a hybrid-attention model, FreeBalance instead profiles the first occurrence of each attention type, such as linear or sparse attention, and reuses the corresponding measurement for later layers of the same type. The planner admits migrations only within
\begin{equation}
B_{\ell}=\max(0,T^{\mathrm{attn}}_{r(\ell)}-\delta),
\end{equation}
where $\delta$ is a safety margin for prediction error and stream interference. The budget is enforced per link and per rank rather than only as a global sum, preventing several individually valid swaps from oversubscribing one participant.


Every rank receives the same globally aggregated count vector and begins from the same placement map. Each rank enumerates candidate pairs, removes conflicts and non-positive-gain candidates, and sorts the remainder by decreasing benefit-to-cost ratio. Ties are resolved lexicographically by layer identifier, source rank, destination rank, and logical expert identifiers. The planner accepts a candidate if neither expert has already been selected and if adding the transfer respects all communication budgets. Loads and residual budgets are updated after each accepted swap.

This deterministic ordering eliminates a plan broadcast: ranks exchange only the compact load statistics required to construct a global view, then reproduce the same swap sequence locally. Determinism also simplifies debugging because a layer, placement map, and predicted count vector uniquely determine the plan. In practice, the planner considers only experts on the most overloaded and most underloaded ranks, which reduces candidate generation from all expert pairs to a small frontier without changing the common heavy-to-light case.

\begin{table*}[t]
\centering
\small
\setlength{\tabcolsep}{4pt}
\renewcommand{\arraystretch}{1.10}
\newcommand{\tbdval}{TBD}
\newcommand{\btbdval}{\textbf{TBD}}

\begin{tabular}{lcccccccc}
\toprule
& \multicolumn{4}{c}{\textbf{Qwen3-30B-A3B-Instruct-2507}}
& \multicolumn{4}{c}{\textbf{Moonlight-16B-A3B-Instruct}} \\
\cmidrule(lr){2-5}\cmidrule(lr){6-9}
\textbf{LongBench Subset}
& \textbf{Vanilla} & \shortstack{\textbf{Vanilla}\\\textbf{+Ours}}
& \textbf{EPLB} & \shortstack{\textbf{EPLB}\\\textbf{+Ours}}
& \textbf{Vanilla} & \shortstack{\textbf{Vanilla}\\\textbf{+Ours}}
& \textbf{EPLB} & \shortstack{\textbf{EPLB}\\\textbf{+Ours}} \\
\midrule
\multicolumn{9}{l}{\textbf{Single-Document QA}} \\
NarrativeQA & 147.6 & \textbf{126.6} & 135.9 & \textbf{126.6} & 77.4 & \textbf{66.7} & 63.6 & \textbf{63.2} \\
Qasper & 102.6 & \textbf{94.6} & 105.6 & \textbf{92.3} & 57.1 & \textbf{55.4} & 57.5 & \textbf{55.3}  \\
MultiFieldQA (EN) & 85.3 & \textbf{74.8} & 76.9 & \textbf{74.7} & 43.1 & \textbf{42.5} & 43.4 & \textbf{42.7}   \\
\multicolumn{9}{l}{\textbf{Multi-Document QA}} \\
HotpotQA & 137.3 & \textbf{122.4} & 121.0 & \textbf{120.9} & 71.6 & \textbf{69.8} & 68.2 & \textbf{67.1} \\
2WikiMQA & 125.2 & \textbf{110.2} & 100.1 & \textbf{93.3} & 66.7 & \textbf{63.1} & 64.9 & \textbf{62.8} \\
MuSiQue & 138.0 & \textbf{123.5} & 122.6 & \textbf{120.1} & 73.0 & \textbf{71.4} & 72.2 & \textbf{69.6} \\
DuReader & 142.4 &\textbf{124.5} & 138.5 & \textbf{123.0} & 78.1 &\textbf{70.7} & 78.7& \textbf{69.5} \\
\multicolumn{9}{l}{\textbf{Summarization}} \\
GovReport & 135.4 & \textbf{118.1} & 132.7 & \textbf{118.3} & 70.0 & \textbf{66.7} & 69.3 & \textbf{66.1} \\
QMSum & 137.6 & \textbf{123.2} & 130.4 & \textbf{122.4} & 73.0 & \textbf{70.8} & 72.7 & \textbf{71.4} \\
MultiNews & 53.6 & \textbf{48.4} & 47.6 & \textbf{47.6} & 27.6 & \textbf{27.3} & 26.9 & \textbf{26.7} \\
VCSUM & 138.1 & \textbf{115.8} & 125.6 & \textbf{112.9} & 73.7 & \textbf{62.6}  & 70.5 & \textbf{61.9} \\
\multicolumn{9}{l}{\textbf{Few-Shot Learning}} \\
TREC & 130.0 & \textbf{118.2} & 118.1 & \textbf{106.8} & 66.5 & \textbf{63.1} & 65.9 & \textbf{62.0} \\
TriviaQA & 175.9 & \textbf{112.1} & 122.6 & \textbf{110.7} & 65.7 & \textbf{64.4} & 64.8 & \textbf{62.7} \\
SAMSum & 131.5 & \textbf{111.8} & 123.1 & \textbf{111.9} & 69.3 & \textbf{62.7} & 66.4 & \textbf{62.8} \\
LSHT & 150.0 & \textbf{124.7} & 151.0 & \textbf{125.3} & 82.6 & \textbf{71.6} & 80.0 & \textbf{69.9} \\
\multicolumn{9}{l}{\textbf{Synthetic Tasks}} \\
PassageCount & 132.2 & \textbf{121.8} & 121.5 & \textbf{121.4} & 72.1 & \textbf{70.0} & 69.9 & \textbf{69.6} \\
PassageRetrieval (EN) & 137.4 & \textbf{123.8} & 136.0 & \textbf{122.7} & 72.1 & \textbf{71.0} & 69.9 & \textbf{69.7} \\
\multicolumn{9}{l}{\textbf{Code Completion}} \\
LCC & 62.3 & \textbf{55.1} & 61.2 & \textbf{54.1} & 37.3 & \textbf{32.9} & 36.9 & \textbf{33.6} \\
RepoBench-P & 129.7 & \textbf{111.8} & 124.7 & \textbf{111.4} & 77.3 & \textbf{67.0} & 77.2 & \textbf{67.9} \\
\multicolumn{9}{l}{\textbf{Mixed-Task Workload}} \\
Mixed Tasks & 83.1 & \textbf{78.0} & 78.3 & \textbf{75.3} & 45.9 & \textbf{43.3} & 44.5 & \textbf{43.1} \\
\bottomrule
\end{tabular}

\caption{Prefill latency across 19 LongBench subsets and one mixed dataset. Columns compare Vanilla, Vanilla with FreeBalance, EPLB, and EPLB with FreeBalance for Qwen3-30B-A3B and Moonlight-16B-A3B. Each entry reports latency in seconds; lower is better.}
\label{tab:end_to_end}
\end{table*}
\section{Evaluation}

\subsubsection*{Models}
We evaluate two MoE models: Qwen3-30B-A3B-Instruct-2507~\cite{yang2025qwen3} and Moonlight-16B-A3B-Instruct~\cite{liu2025muon}. Qwen3-30B-A3B-Instruct-2507 contains 128 experts, activates the top-8 experts per token, and places 16 experts on each EP rank; each expert occupies 9\,MB. Moonlight-16B-A3B-Instruct contains 64 experts, activates the top-6 experts per token, and places 8 experts on each EP rank; each expert occupies 16.5\,MB. In both models, one expert constitutes the basic unit of expert migration.

\subsubsection*{Benchmarks}
We use LongBench~\cite{bai2024longbench} as the primary workload suite because it covers six distinct long-context capabilities. Table~\ref{tab:end_to_end} reports 19 subsets: covering single- and multi-document QA, summarization, few-shot learning, synthetic tasks, and code completion. Each subset is evaluated independently under the same batching and sequence-length configuration.

To emulate dynamic multi-task serving, we additionally construct the Mixed Tasks workload as a sequence of prefill steps whose LongBench subset changes from one step to the next. Each individual prefill step draws its requests from one subset. The workload therefore captures step-to-step shifts in routing distributions while retaining the same batch size and input-length configuration as the subset-specific runs. Its row reports latency under this changing-subset execution sequence. For the history-driven EPLB baseline, we construct its profiling state separately by aggregating expert-load statistics from 20 samples from each of all 21 LongBench subsets, for 420 profiling samples in total.

\subsubsection*{Configurations}
Experiments run on nodes equipped with 8$\times$ NVIDIA A800-SXM4 GPUs connected by NVLink. We use expert parallelism across all eight GPUs (EP$=8$). Unless stated otherwise, we use a batch size of 16 with an input length of 8K tokens. Each configuration is warmed up once and then measured for three runs; we report the average of the measured runs. We compare Vanilla with a fixed expert placement, Vanilla with FreeBalance, history-driven EPLB, and EPLB with FreeBalance under the same model weights, routing decisions, batching policies, and parallel configurations.

\subsubsection*{Metrics}
We report end-to-end prefill latency and the max-to-mean rank-load ratio, where a value of one indicates perfect balance. For residual pre-routing, we measure the cosine similarity between $h_{\ell-1}$ and $H_{\ell}$, the cosine similarity between their router logits, and the top-$k$ hit rate, defined as the fraction of final top-$k$ assignments recovered by the early routing pass. Planner quality is measured by the fraction of executed swaps that reduce the realized imbalance. All GPU timings are collected with synchronized CUDA events.

\subsection{End-to-End Effectiveness}
\label{subsec:end_to_end}
Vanilla reaches per-layer max/mean ratios up to $2.01$, whereas FreeBalance reduces them to $1.35$, an improvement of up to $32.8\%$. The gain spans communication and computation: a more even token distribution reduces both the maximum per-rank activation volume during dispatch and combine and the maximum per-rank expert workload. Since all-to-all collectives and expert execution are gated by the slowest EP rank, mitigating this straggler accelerates the entire MoE stage.

Across the 19 evaluated LongBench subsets, FreeBalance reduces Qwen3-30B's average prefill latency by $13.1\%$. Because its auxiliary work overlaps with existing critical-path computation, as analyzed in Section~\ref{subsec:overlap_analysis}, online reconfiguration introduces no separate pause. The load-balance improvement does not translate linearly into wall-clock speedup because other kernels remain on the critical path. 

\subsection{Prediction and Planning Quality}

Table~\ref{tab:prerouting_quality} compares the early router invocation $g_{\ell}(h_{\ell-1})$ with the normal invocation $g_{\ell}(H_{\ell})$. Across both models and all three datasets, logit cosine similarity remains between $0.9896$ and $0.9952$, even though hidden-state cosine similarity ranges from $0.7116$ to $0.9316$. The resulting top-$k$ hit rate ranges from $0.7419$ to $0.8494$. These results show that the router logits remain strongly aligned across the two invocation points and provide a useful signal for estimating the target layer's expert workload.

Across layers in which migrations are executed, $95\%$ achieve a lower measured max-to-mean rank-load ratio after applying the complete swap plan. This rate is measured at the layer level rather than for individual swaps. The lack of improvement in the remaining layers arises from runtime variation or interactions among multiple swaps whose isolated benefits are not perfectly additive.

\subsection{Overlap Analysis}
\label{subsec:overlap_analysis}

Residual pre-routing and planning consume computation, whereas the measured expert migrations require approximately $3$--$4$\,ms per layer. A sequential design would add these costs directly to prefill latency. In the two-stream schedule, the early router invocation and planning execute concurrently with the target layer's attention. Once the plan is available, expert-weight transfer overlaps with the remaining attention computation. The remaining question is how many swaps can be admitted without exposing a migration tail on the critical path.

\begin{table}[t]
\centering
\small
\setlength{\tabcolsep}{2.5pt}
\begin{tabular}{llccc}
\toprule
Model & Dataset & Hidden cos. & Logit cos.  & Top-$k$ hit \\
\midrule
\multirow{3}{*}{Qwen}
  & PassageRetrieval & 0.7334 & 0.9952 & 0.7520 \\
  & LSHT     & 0.7242 & 0.9951 & 0.7578 \\
  & TriviaQA & 0.7116 & 0.9947 & 0.7419 \\
\midrule
\multirow{3}{*}{Moonlight}
  & PassageRetrieval & 0.9316 & 0.9900 & 0.8461 \\
  & LSHT     & 0.9291 & 0.9896  & 0.8494 \\
  & TriviaQA & 0.9234 & 0.9910  & 0.8037 \\
\bottomrule
\end{tabular}
\small
\caption{Pre-routing quality across models and datasets. Router logit and hidden cosine similarities compare the early and normal routing passes; top-$k$ hit measures assignment overlap.}
\label{tab:prerouting_quality}
\end{table}

\begin{figure}[t]
\centering
\includegraphics[width=\columnwidth]{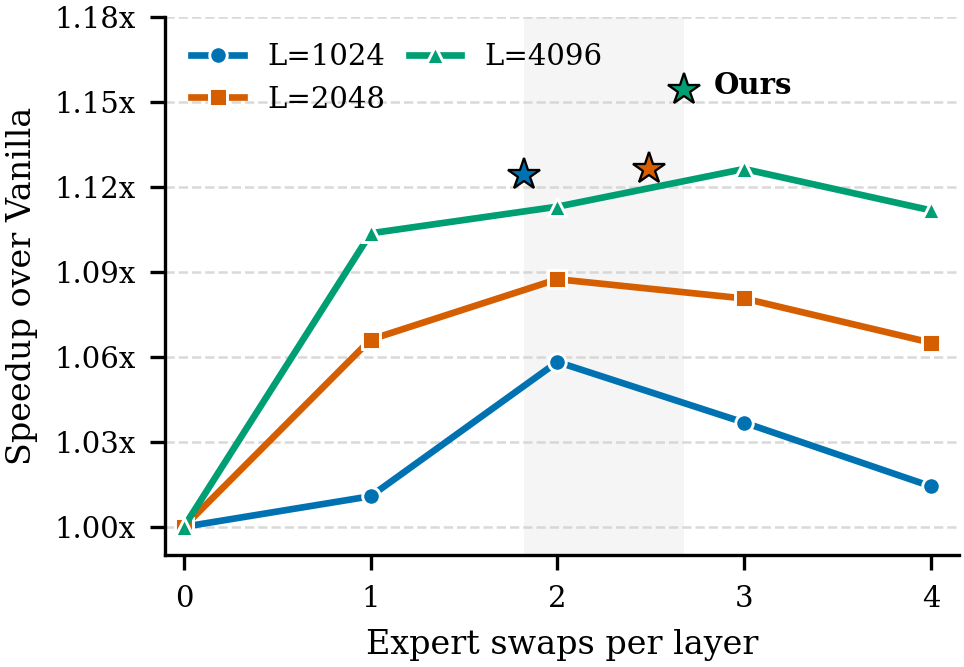}
\caption{Speedup over Vanilla as a function of expert swaps per layer. Lines show zero to four forced swaps per layer; stars show adaptive FreeBalance, whose average swap count ranges from 1.82 to 2.68 across sequence lengths.}
\label{fig:ttft_vs_swaps}
\end{figure}

Figure~\ref{fig:ttft_vs_swaps} shows that speedup is not monotonic in a fixed number of swaps per layer. The best fixed policies achieve $1.0582\times$, $1.0874\times$, and $1.1263\times$ speedups at 1K, 2K, and 4K tokens, respectively. By adapting the average swap count from 1.82 to 2.68, FreeBalance reaches $1.1241\times$, $1.1264\times$, and $1.1542\times$, outperforming the fixed policies at all three lengths. These results show that the overlap budget should determine the migration count rather than a fixed policy. 

\subsection{Sensitivity to Sequence Length}

We evaluate how sequence length affects the balancing benefit and the opportunity to hide expert migration. We fix the batch size to 16 and vary the input length from 1K to 8K tokens. For each configuration, we report time to first token (TTFT) and the max-to-mean rank-load ratio for Vanilla and FreeBalance. We additionally report the measured overlap time available for migration.

\begin{table}[t]
\centering
\small
\setlength{\tabcolsep}{1.8pt}
\renewcommand{\arraystretch}{1.08}
\begin{tabular*}{\columnwidth}{@{\extracolsep{\fill}}lccccc@{}}
\toprule
\multirow{2}{*}{Length}
  & \multicolumn{2}{c}{Vanilla}
  & \multicolumn{2}{c}{Ours}
  & \multirow{2}{*}{\shortstack{Overlap\\(ms)}} \\
\cmidrule(lr){2-3}\cmidrule(lr){4-5}
  & TTFT & Max/Mean & TTFT & Max/Mean & \\
\midrule
1K & 3.08  & 2.00 & \textbf{2.74}  & \textbf{1.33} & 12.55 \\
2K & 4.99  & 2.02 & \textbf{4.43}  & \textbf{1.34} & 17.35 \\
4K & 8.98  & 2.03 & \textbf{7.78}  & \textbf{1.35} & 31.30 \\
8K & 20.45 & 2.04 & \textbf{15.68} & \textbf{1.37} & 36.80 \\
\bottomrule
\end{tabular*}
\caption{Performance under batch size 16. TTFT is reported in seconds, and overlap time is reported in milliseconds.}
\label{tab:batch16_scaling}
\end{table}

As shown in Table~\ref{tab:batch16_scaling}, FreeBalance reduces TTFT by $14.7\%$ on average across the evaluated sequence lengths. The reduction grows from $11.0\%$ at 1K tokens to $23.3\%$ at 8K tokens. Meanwhile, the max-to-mean ratio decreases from a Vanilla range of $2.00$--$2.04$ to a FreeBalance range of $1.33$--$1.37$. The measured overlap time increases from $12.55$\,ms to $36.80$\,ms, indicating that longer sequences provide a larger window for hiding expert migration.




\section{Related Work}
\noindent\textbf{Offline balancing.}
DeepSeek EPLB periodically replicates and places experts using historical routing statistics \cite{deepseek2025eplb}. While effective for slowly varying workloads by amortizing optimization and weight rearrangement, its historical workload estimation may lag behind abrupt task shifts and cannot directly correct layer- and batch-specific imbalance, especially in multi-task serving where consecutive batches activate different expert subsets. FreeBalance instead predicts the upcoming workload within the current prefill and adjusts placement at layer granularity.

\noindent\textbf{Online balancing.}
Dynamic systems adapt expert placement to observed demand \cite{he2022fastermoe,li2023accelerating}, while Harmony coordinates runtime scheduling with model-state movement \cite{li2022harmony}. Recent approaches such as UltraEP and MoonEP rebalance the realized routing distribution using redundant experts, with runtime planning and expert-state movement after routing decisions \cite{wei2026ultraep,chen2026moonep}. Such reactive strategies depend on already-observed routing patterns and provide limited opportunity to hide rebalancing overhead before expert computation. FreeBalance instead anticipates upcoming expert demand before the target routing stage, budgets swaps within the available attention window, and overlaps expert migration with main-stream computation.

\section{Conclusion}
FreeBalance addresses dynamic load imbalance in distributed MoE inference by anticipating expert workloads and initiating balancing before target-layer dispatch. By shifting balancing earlier, it adapts to each workload while hiding reconfiguration overhead from the critical path and preserving the original routing decisions and outputs. Experiments show that FreeBalance reduces the max-to-mean rank-load ratio by up to 32.8\% and end-to-end prefill latency by 13.1\%.

\clearpage
\bibliography{aaai2027}
\end{document}